\documentclass[sigconf, 10pt]{acmart}

\usepackage{booktabs} 
\usepackage{soul}
\usepackage{graphicx} 
\usepackage{caption,subcaption}
\usepackage{hyperref}
\usepackage{natbib}
  {%
      \end{minipage}%
        \end{center}}
\definecolor{mintgreen}{rgb}{0.6, 1.0, 0.6}
\definecolor{pastelviolet}{rgb}{0.8, 0.6, 0.79}
\definecolor{peridot}{rgb}{0.9, 0.89, 0.0}
\definecolor{richbrilliantlavender}{rgb}{0.95, 0.65, 1.0}
\definecolor{robineggblue}{rgb}{0.0, 0.8, 0.8}

\setcopyright{none}

\begin{document}
\title{Enabling Collaborative Video Sensing at the Edge through Convolutional Sharing}

\author{Kasthuri Jayarajah, Dhanuja Wanniarachchige, Archan Misra}
\affiliation{%
  \institution{Singapore Management University}
}
\email{kasthurij,dhanujaw,archanm@smu.edu.sg}

\begin{abstract}
While Deep Neural Network (DNN) models have provided remarkable advances in machine vision capabilities, their high computational complexity and model sizes present a formidable roadblock to deployment in AIoT-based sensing applications. In this paper, we propose a novel paradigm by which peer nodes in a network can collaborate to improve their accuracy on person detection, an exemplar machine vision task. The proposed methodology requires no re-training of the DNNs and incurs minimal processing latency as it extracts scene summaries from the collaborators and injects back into DNNs of the reference cameras, on-the-fly. Early results show promise with improvements in recall as high as 10\% with a single collaborator, on benchmark datasets.
\end{abstract}

\maketitle

%
%


\section{Introduction}
\label{sec:intro}

Supporting deep learning based \emph{machine intelligence} for vision tasks on resource-constrained embedded devices is currently one of the hottest research challenges. Such intelligence can enables a range of novel applications across mobile/wearable devices and IoT devices (e.g., visual tracking using vehicle-mounted cameras~\cite{ chen2015glimpse}). The key challenge of course is the mismatch between the prohibitive computational load of DNN (Deep Neural Network) pipelines and the resource constraints (memory, GPU, energy) of such wearable and IoT devices. This mismatch continues to increase in severity, especially as the deep learning community continues its push for improving visual perceptual accuracy at the expense of increased computational complexity. Interestingly, recent trends in DNN research suggest an emerging regime of ``diminishing returns", where state-of-the-art ``deeper" DNNs require \emph{increasingly higher depth (and thus, a significantly higher computational overhead) to achieve modest increases in accuracy}. Witness, for example, the emergence in recent years of YoLo v3~\cite{YOLO3} (a 106 layer-deep DNN model with $\sim$53 convolutional layers) in contrast to the Vgg16-SSD~\cite{liu2016ssd} (a 23 layer-deep model): while YoLo v3's processing throughput is less than half that of SSD, it's accuracy gain on benchmark datasets~\cite{ferryman2009pets2009} is comparatively lower. 

\begin{figure}[tbh]
\begin{center}
\includegraphics[width=0.75\linewidth,height=1.2in]{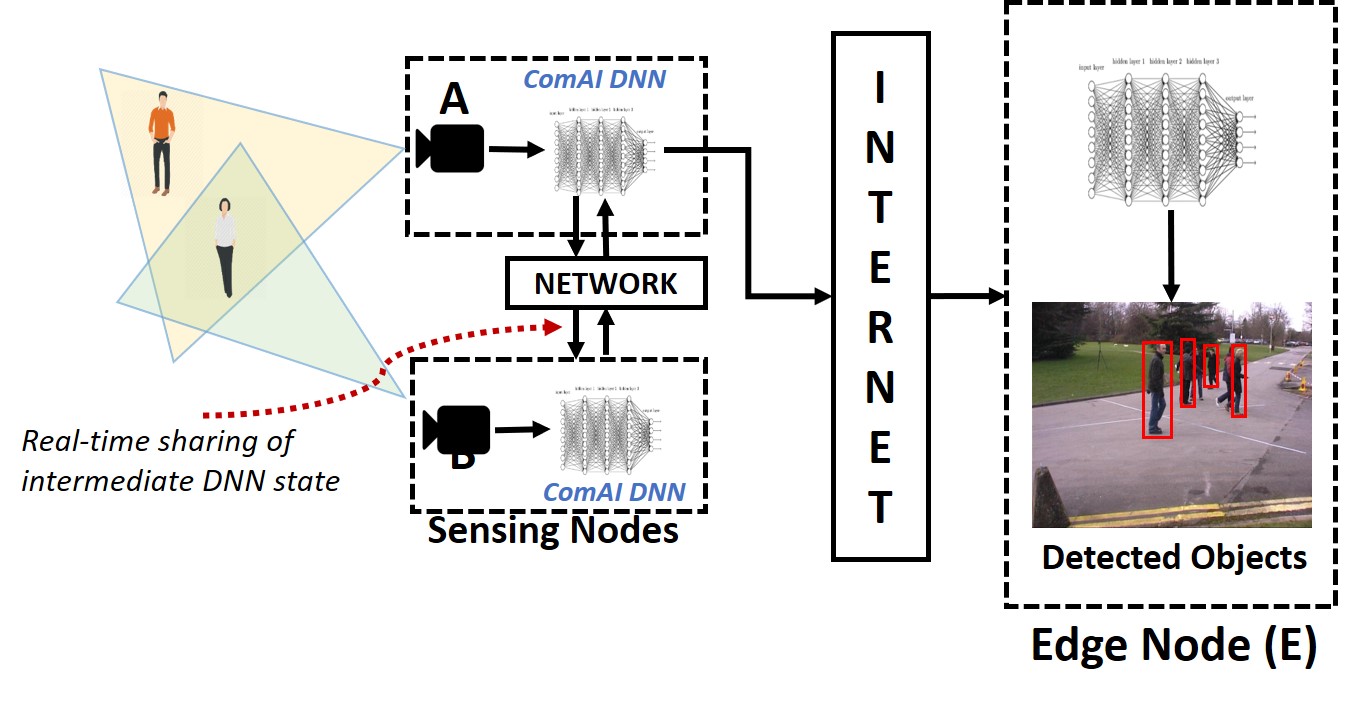}
\vspace{-0.15in}
    \caption{The Collaborative (\& Networked) Video Sensing Paradigm.}
    \vspace{-0.1in}
\label{fig:collabarch}
\end{center}
\end{figure}

Our work is motivated by two key recent trends: (a) high-density deployments where IoT nodes collectively sense \emph{correlated} observations of the same physical phenomena, and (b) dramatic reduction in energy per transmitted bit, making it possible to exchange larger volumes of data with \emph{nearby} battery-operated nodes. In this work, we seek to explore the fundamental question: \emph{Can collaborative nodes exchange summaries of their perspective of a sensed phenomena, of low complexity (faster but less accurate) DNNs, to improve their sensing accuracy to match that of higher complexity models?} To this end, we propose a paradigm for collaborative machine intelligence, and present our early efforts in enabling collaboration in vision sensing, in particular, we consider the person detection task where camera nodes share scene summaries in the form of feature maps (i.e., selected outputs from the early, convolutional layers).

\noindent \textbf{Key Contributions:} Through this work, we make the following key contributions:
\begin{enumerate}
    \item{\emph{\textbf{Identifying Causes for Misclassification and Opportunities for Collaboration}}: We take a state-of-the-art object detector as an exemplar, dissect it at intermediate stages of the DNN pipeline to characterize common issues that lead to inaccuracies in the detection process. Specifically, we draw comparisons of (1) the raw activation values of feature maps at the initial convolutional layers, and (2) confidence values at the output layers, between correct and incorrect detections to identify opportunities for peers in a network, overlapping fields of view (FoV) but different perspectives, to \emph{collaborate} to overcome such errors.}
    \item{\emph{\textbf{Improving DNN Accuracy through Feature Manipulation}}: We propose an workflow for collaboration through convolutional sharing and feature manipulation. Here, nodes share summaries of their perspective as early as the initial convolutional layers, with their peers, which are then suitably fed back into the later layers of the DNN for improved accuracy. Such run-time adaptation of the network incurs no model re-training, minimal processing latency overhead and gracefully rolls back to the standalone mode when collaborative input is not available.}
    
\end{enumerate}

\section{Opportunities for Collaboration in Video Sensing}
\label{sec:opp}
In this paper, we take an exemplar scenario *(illustrated in Figure~\ref{fig:collabarch}) of multiple networked vision sensors (video cameras), deployed in campus-like settings, and demonstrate the benefits of collaborative machine intelligence for a typical \emph{vision-based human/object detection and localization} task. We emphasize at the outset that this conceptual framework is merely an exemplar---while the detailed mechanics of state-sharing and evaluation results presented here are based on a multi-camera instance, the concepts apply to a broader range of AIoT inferencing pipelines. 

\subsection{Dissecting the DNN Pipeline}
In this subsection, we dissect the different stages of a state-of-the-art object detection DNN (i.e., SSD~\cite{liu2016ssd}) to identify problems that lead to errors. A correct detection is one where a bounding box is output by the DNN correctly identifies the object class, and the box is localized with a certain amount of intersection with a ground-truth bounding box. SSD, as well as many other single shot detectors, rely on the concept of anchor boxes -- these serve as a discretization of the image space where class confidences are estimated at the granularity of such boxes, and the same boxes are scaled/translated to output the final, localized bounding box detections. We use sample frames from the PETS2009~\footnote{\url{http://www.cvg.reading.ac.uk/PETS2009/a.html}} in the analyses that follow.

\begin{figure}[htbp]
  \begin{minipage}[htbp]{0.23\textwidth}
    \includegraphics[width=0.95\linewidth,height=1.1in]{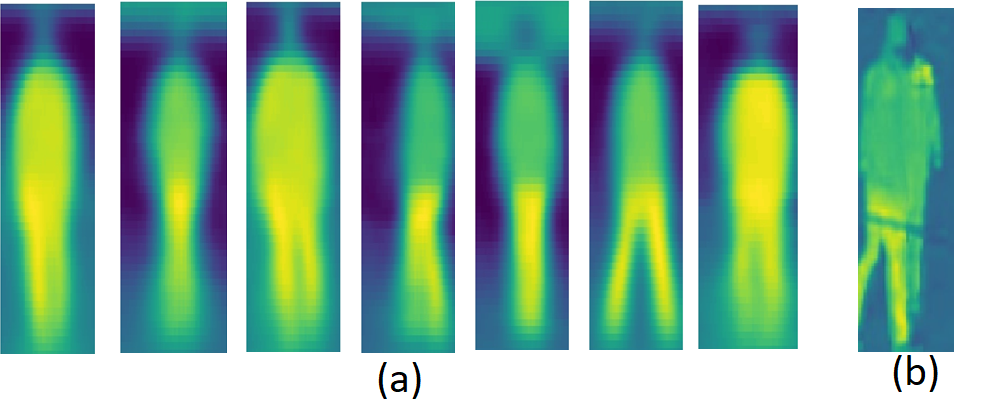}
     \vspace{-0.2in}
    \caption{Examples of fmap output for (a) correct detections and (b) an incorrect detection due to overlapping person objects in the scene.}
\label{fig:clusters}
  \end{minipage}
  \hfill
  \begin{minipage}[htbp]{0.23\textwidth}
    \includegraphics[width=1.6in,height=1.1in]{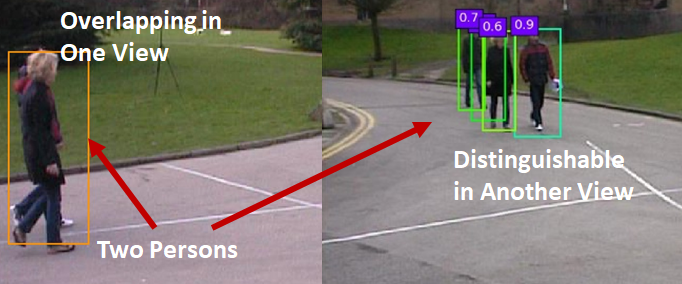}
    \vspace{-0.2in}
    \caption{Distance Ambiguity of Person Objects across Two Camera Views}
    \label{fig:overlapping}
  \end{minipage}
  \vspace{-0.2in}
\end{figure}

\textbf{Feature Layers: } A key observation we make from analyzing \emph{feature maps} (fmaps), i.e., outputs of filters from convolutional layers of the DNN, is that if the distribution of the raw activation values (corresponding to the person object) differs immensely from the distributions seen during the training phase, the DNN is likely to filter \emph{out} the person in the subsequent stages of the pipeline. This situation can arise when there's a partial or occluded view of the person or when the person is in a position that is not distinguishable from other object classes. In Figure~\ref{fig:clusters}, we show \emph{templates} of activations that were correctly identified (clustered with $k-means$ clustering) against an example of a person that was not detected due to occlusion.


\begin{figure}[tbh]
\centerline{\includegraphics[width=0.3\textwidth]{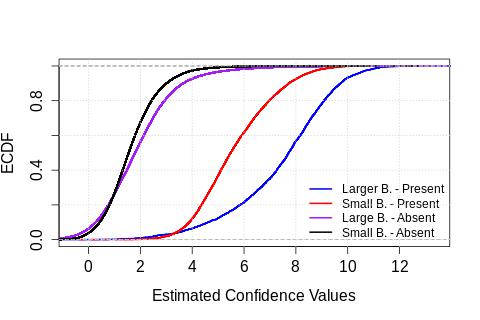}}
\caption{Comparison of estimated Person class confidence values for small and large anchor boxes.}

\label{fig:conf-values}
\vspace{-0.3in}
\end{figure}

\textbf{Output/Predictor Layers: }Even if a person object's presence (via suitably high activation values) is retained by the convolutional layers, the class confidence assigned to the object, at the later \emph{predictor} layer, can still be low. In Figure~\ref{fig:conf-values}, we plot the CDF of estimated confidence values by the DNN, for the ``person" class, for when a person was present within the anchor box vs. not, further separated by the size of the anchor box. We label anchor boxes corresponding to the first predictor layer to be ``small" (that detect smaller objects) and the remaining as ``large" (that detect larger objects). The SSD architecture consists of 6 predictor layers. We highlight to the reader, that there exists a significant difference between the confidence values for small (red line) and large boxes (box line) resulting in ``smaller" person objects being less likely to be detected by the DNN.


\subsection{Opportunities for Collaboration}
Based on the above observations, we identify the following opportunities where collaborative input can help alleviate issues leading to misdetections.

\textbf{Convolutional Sharing and Feature Manipulation: }Issues such as occlusion and mismatch against training samples can cause early feature layers of the pipeline to completely disregard a person object falsely. We posit that by sharing fmaps, from collaborators with \emph{better} perspectives, due to the sheer difference in their positioning, a reference camera can in fact ``augment" its view, at the feature map level. In the following sections of the paper, we propose an architecture and prescribe a methodology for extracting convolutional information from peers and injecting back into the DNN execution, for improving the detection accuracy.

\textbf{Confidence Manipulation: }While a DNN can misclassify a person as non-person due to reasons such as its size (smaller objects are les likely to be detected), the same person, could appear as a larger object in a collaborator's view, due to the person being more proximal. As a result, a DNN could learn from its peers to override or boost its estimated class confidence probabilities. In our recent work~\cite{comai2021}, we propose and evaluate a methodology for manipulating predictor layer outputs for improving accuracy. 

\textbf{Localization Improvement: }The non-uniform distribution of people across a scene calls for non-uniform measures to correctly \emph{distinguish} person objects at the output stage (e.g., through measures such as adaptive thresholds at the NMS stage). A key opportunity to exploit here is the fact that two persons overlapping in one scene can in fact be seen far apart, with no or minimal overlap in a collaborator's view, as seen in the example in Figure~\ref{fig:overlapping}. Hence, a NMS algorithm designed to transfer the knowledge of distance ambiguity across collaborating views can be useful in improving the detection accuracy.
\vspace{-0.3in}

\section{System Overview}
\label{sec:system}
Based on our observations in Section~\ref{sec:opp}, we propose an architecture for collaborative, peer-to-peer sharing and fusion (as illustrated in Fig~\ref{fig:sysoverview}). In  Figure~\ref{fig:collab-operation}, we propose an early workflow for \emph{on-the-fly} adaptation of the DNNs, for extracting scene summaries and then feeding back collaborative input for accuracy gains. 

We emphasise that a key design consideration for our collaborative paradigm is that it should not require any \emph{re-training of the DNN} which requires both a large number of training frames and manual labeling effort. The ability to adapt at run-time also allows for seamless rolling back to the standard, independent execution mode when collaborative input is not available. We make the key assumption that the nodes are aware of the perspective transformation (e.g., through homography between them and other nodes).  

\subsection{System Components}
\begin{figure}[t]
\centerline{\includegraphics[width=0.45\textwidth, height=1.5in]{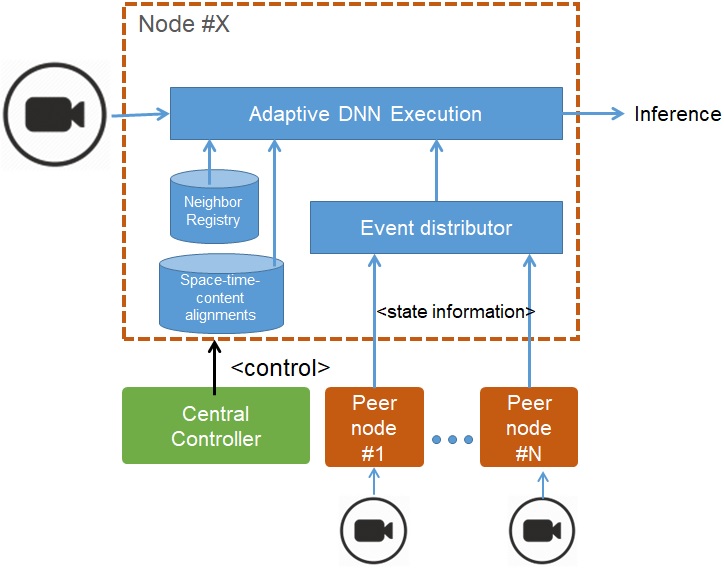}}
\vspace{-0.1in}
\caption{Proposed Architecture for Collaborative Sensing on the Edge}
\vspace{-0.35in}
\label{fig:sysoverview}
\end{figure}

In Figure~\ref{fig:sysoverview}, we illustrate our proposed system for collaborative vision sensing. The architecture involves a central control node and a network of peer cameras. Node $X$ refers to a generic peer node which receives control commands from the Control Center and state information (e.g., inferred bounding boxes, statistical features, etc.) from other peer nodes in the network. 

\textbf{Central Controller: }This module acts as the broker between peer camera nodes. The controller could be a camera node itself, or a more resource-rich device such as a cloudlet~\cite{satyanarayanan2017edge}. The module receives inference results and state information from the other peer nodes to perform several functions. It is also in charge of setting and updating the mode and various other configurations of the network.

\textbf{Peer Nodes: }We assume that each peer node has processing capabilities on-board (e.g., CPU or a mobile Vision Processing Unit (VPU)) that allows them to execute full, or partial, DNN pipelines. The peer nodes directly exchange send/receive state information to/from chosen peers. Depending on the mode of operation, the nodes use the information from peers as input, in addition to the node's own view of the scene, to run collaborative DNN pipelines. The nodes consist of two data stores: (1) the Neighbor Registry which contains look-up information on local addresses of collaborating peers, additional information such as the peers' reputation scores~\cite{fusion19}, etc. and (2) the Spatial Mapper which consists of information key to translating inferences of a peer's view to a node's own view. The Event Distributor in the node acts as a scheduling service -- for instance, for a person predicted to arrive at a future time instant (using peer inference), the distributor marks the relevant information such that the node uses it only at the appropriate moment in future. Although our current work assumes only learning from concurrent frames amongst the peers, future work will extend it to such non-synchronous situations.

\subsection{Collaborative Workflow}
\begin{figure}[t]
\centerline{\includegraphics[width=0.3\textwidth]{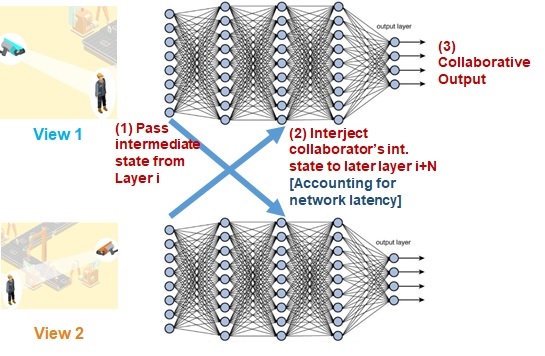}}
\caption{Overview of Collaborative Operation for Improved Accuracy}
\vspace{-0.2in}
\label{fig:collab-operation}
\end{figure}

In this work, we propose a collaborative workflow for improving accuracy through \textbf{convolutional sharing and feature manipulation }(as illustrated in Figure~\ref{fig:collab-operation}). 

For simplicity, we will refer to collaborative nodes as those nodes that share scene summaries, and reference nodes as those that ingest summaries from collaborators. However, we point to the reader, that under the collaborative paradigm, \emph{all} nodes operate under both the reference and collaborator modes.

\textbf{Step 1: }As the initial convolutional layers finish execution, a summary/digest of those intermediate outputs are extracted and shared over the network with \emph{chosen} peers in the network, \emph{in parallel}.

\textbf{Step 2: }The digests reach the reference nodes with a time delay (due to the transmission over the network). As such, the digests are then appropriately transformed to be ingested by the reference node's DNN, that is now executing at a later (convolutional) layer.  

\textbf{Step 3: }The DNN at the reference node, now with the fusion of collaborative input form one or more collaborators, completes execution as per normal.

We highlight to the reader that the nodes incur additional processing latency due to the scene summary extraction from an early layer and the ingestion at a later layer (both of which are insignificant in relation to the total latency of the DNN execution), but do not incur any network latency due to the fact that the DNNs do not \emph{wait} on input from collaborators. In the following section, we describe the details of Steps 1 and 2.

\section{Early Layer Feature Exchange for Collaborative Video Sensing}
\label{sec:feature}

\begin{figure}[t]
\centerline{\includegraphics[width=0.35\textwidth,height=2in]{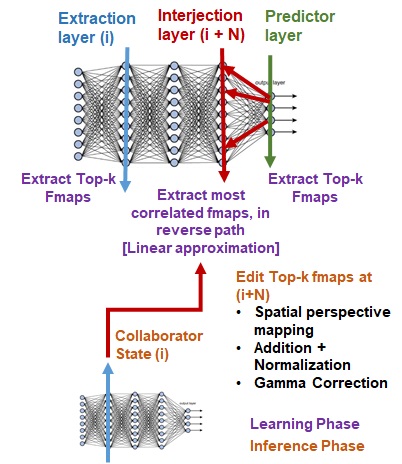}}
\caption{Steps during the Learning and Inference Phases of the Collaborative Operation}
\vspace{-0.2in}
\label{fig:collab-acc-steps}
\end{figure}

As described before, the edge nodes run the default DNN pipelines independently with (Step 1:) scene summary/digest extraction and (Step 2:) run-time interception of the DNN to ingest collaborative input. 
As the digest is broadcast over the network, it is required that the digest is small in size and yet summarize the scene observed the peer legibly. As sharing over a network incurs a time delay, it is preferred that the digest extraction occurs as early as possible in the pipeline whilst the incorporation of peer information happens as later in the pipeline as possible. We describe a methodology for extracting scene summaries in our previous work~\cite{comai2021, kasthesis}. The output from Step 1 is a summary (e.g., average) of set of systematically chosen \emph{feature maps (fmaps)}, i.e., output filters of convolutional layers, that best represent the objects of interest (persons, in this case). 

In Section~\ref{sec:intercept} we elaborate the processes for learning \emph{where} to intercept and \emph{how} to intercept, for improved accuracy.

\subsection{Intercepting DNNs with Peer Information}
\label{sec:intercept}
After receiving the summary fmaps from collaborators, the camera nodes also need to learn at \emph{what} depth of the DNN and \emph{which} filter outputs at that depth should they augment to  improve their detection performance. We  describe the steps of our technique to derive which filter outputs can be augmented (and leave the determination of the best depth to collaborate for future work):
\begin{enumerate}
\item{For each predictor layer, occurring later in the pipeline, we extract the top-$k_n$ discriminant filters using the same method described to extract scene summaries.}
\item{Then for each such filter, we trace back in the reverse direction, at each $i+N$ convolutional layer for the most correlated filters. Correlation assumes a linear approximation between the filter outputs between the layers although the activation units are essentially \emph{non-linear}. For simplicity, we assume single order regressions.}
\item{Then, during run-time, the top-$k'$ filters at the $i + N^{th}$ layer are augmented with the collaborators' input as described below.}
\end{enumerate}

Here, the $i^{th}$ layer refers to the layer where scene summaries were extracted from. During the run-time, once the (early layer) digests from collaborator cameras are received, the activation values from the peers (in the form of fmaps) are introduced to the activation values of the reference camera's own activation values additively, after being spatially mapped. Such addition must also be re-scaled appropriately to match the smaller size of the filers at the later stages. Following the perturbation step, the activation values are re-normalized and Gamma corrected. While the re-normalization step helps in ``boosting" the pixels the collaborators affirm that those are indeed pixels related to the target (e.g., person), the Gamma correction helps in retaining the activations of the pixels detected as persons in the camera's own scene but are not affirmed by any other camera due to no existing overlaps. 

\section{Early Results}
\label{sec:eval}
In this section, we show early results from our proposed methodology for feature manipulation, on-the-fly, for (a) improving person detection accuracy and (b) the feasibility of sharing and fusing collaborator input given network delays.
\vspace{-0.2in}
\subsection{Person Detection Performance under Realistic Settings}
In this section, we evaluate the improvement in accuracy the standard DNNs are able to gain through collaboration with peers in the network. To this end, we evaluate the person detection accuracy across the different reference views from both the PETS~\cite{ferryman2009pets2009} and WILDTRACK~\cite{wildtrack:Chavdarova_2018_CVPR} datasets. Here, we use the SSD object detector, with scene summaries extracted from layer $i=1$, i.e., the first convolutional layer, and ingest back at the layer immediately preceding the first predictor layer, i.e.,$i+N=9$.

In Table~\ref{tab:wt-pair-benchmark}, we report accuracy metrics (measured as precision, recall and F-score of detection averaged over all frames), as percentages, for the baseline scenario of where the reference camera runs the standard DNN pipeline with no input other than its own view, against augmented input (intercepted at some later layer in the pipeline) from collaborators. We consider the addition of collaborators in the order of increasing amount of overlap with the reference view. We provide the accuracy numbers for both intercepting at a convolutional layer ((a) with activation values transferred from peers vs. (b) binary values indicating the locations within the view to augment) as well as intercepting at predictive layers (overriding object detection confidence values). 

We make the following observations:
\begin{enumerate}
\item{The gain in detection accuracy is significant even with as few as 1 or 2 collaborators -- for e.g., in Table~\ref{tab:wt-pair-benchmark}, we see that the precision and recall improve by 8\% and 4\%, respectively. }
\item{Note that sharing locations where activation values should be boosted ($C_l$) is more favorable to sharing the fmaps themselves ($C_a$). This implies that the digests transferred over the network could be even smaller. }
\item{We also note in Table~\ref{tab:pets-all-gamma}, similar performance gains are observable across datasets -- the recall improves by as much as 10\% on the PETS dataset.}
\end{enumerate}

We highlight to the reader that the area of overlap and thus the amount of ``information gain" from collaboration is constrained by the deployment design. 

\begin{table}[]
\resizebox{\linewidth}{!}{
\begin{tabular}{p{2cm}p{1cm}p{1cm}p{1cm}p{1cm}p{1cm}p{1cm}p{1cm}}
\hline
$R$= 4           & Baseline & $C_a$=1 & $C_l$=1 & $C$=7 & $C$=1,7 & $C$=all \\ \hline
Precision        & 21.11    & 24.03                     & 26.81             & 26.72 & 27.75   & 29.29           \\
Recall           & 21.93    & 25.21                     & 25.36             & 25.36 & 25.37   & 25.69           \\
F-score          & 21.51    & 24.62                     & 26.06             & 26.02 & 26.50   & 27.37           \\
Precision (Gain) &          & 13.83                     & 27.00             & 26.58 & 31.43   & 38.74           \\
Recall (Gain)    &          & 14.95                     & 15.64             & 15.64 & 15.67   & 17.16           \\
F-score (Gain)   &          & 14.44                     & 21.15             & 20.97 & 23.21   & 27.26    \\\hline      
\end{tabular}
}
\caption{Illustrative Person Detection Performance for a Reference Camera 4 in the WILDTRACK dataset with and without collaboration. $C_a$ -- both positions and activation values shared, and $C_l$ -- only positions shared.}
\label{tab:wt-pair-benchmark}
\vspace{-0.3in}
\end{table}


\begin{table}[]
\resizebox{\linewidth}{!}{
\begin{tabular}{llll}
\hline
 & \textbf{Precision}                      & \textbf{Recall}  & \textbf{F-Score}  \\ \hline
$R$=7 (Baseline)               & 95\%    & 56.98\% & 71.44\% \\
$C_l=all$ (normalization only) & 20\%    & 70\%    & 33\%    \\
$C_l=1$ (w/Gamma)              & 94\%    & 66\%    & 76\%    \\
$C_l=5$ (w/Gamma)              & 94.24\% & 64.20\% & 76.38\% \\
$C_l=6$ (w/Gamma)              & 93\%    & 66\%    & 76\%    \\
$C_l=8$ (w/Gamma)              & 94.16\% & 63.39\% & 75.77\% \\
$C_l=all$ (w/Gamma)            & 93.77\% & 65.12\% & 76.86\% \\ \hline
\end{tabular}
}
\caption{Person Detection Performance with on the PETS dataset.}
\vspace{-0.3in}
\label{tab:pets-all-gamma}
\end{table}

\begin{figure}[tbh]
\begin{center}
\includegraphics[width=0.65\linewidth]{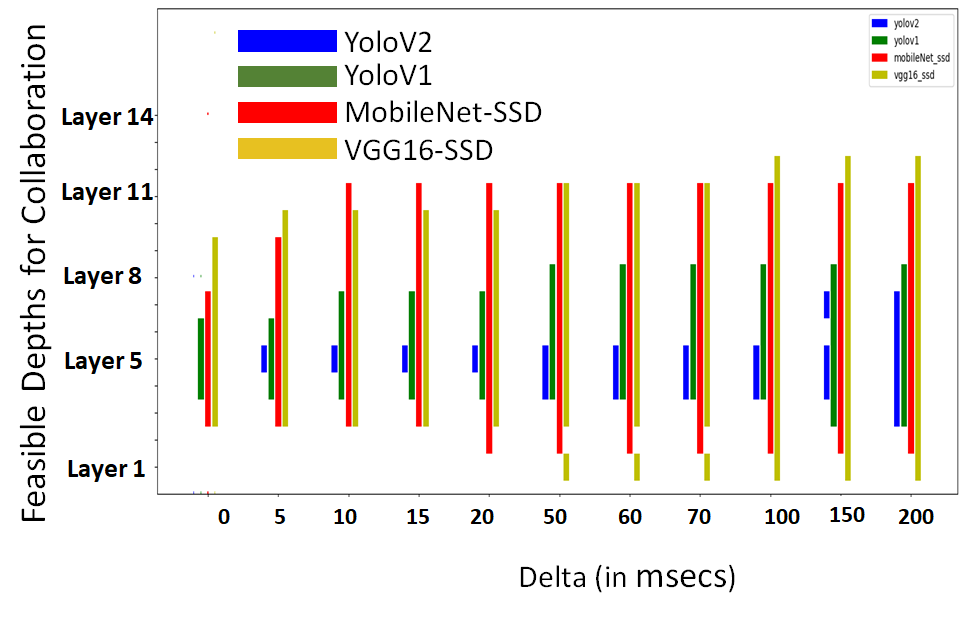}
    \caption{Feasibility of digest sharing at different depths, for multiple state-of-the-art object detector DNNs.}
\label{fig:final-tradeoff}
\end{center}
\vspace{-0.3in}
\end{figure}

\subsection{Feasibility of Collaboration over the Network}
Next, we quantify the feasibility of extracting and transferring digests while accounting for network delays\footnote{Nodes running on Jetson Nano boards, with Raspberry Pi Cameras V2, connected over a private WiFi network.}.

In Figure~\ref{fig:final-tradeoff}, we show the pairs of consecutive layers for which collaboration is feasible -- i.e., digest extracted from layer $i$ and ingested back at layer $i+1$. The $x-$ axis shows the time bound $\Delta$ (in msecs) within collaboration is feasible such that $computeTime(i+1)+ WiFiTransmitTime(i+1)< \Delta + computeTime (i)$. We observe that even with $\Delta=0$ (i.e., reference camera doesn't \emph{wait}), there are pairs of layers for which on-the-fly collaboration is still feasible. We point out that this is a stringent condition (collaborating on consecutive layers) and that as we show in the previous results, collaborating on farther apart layers (between the 1st and 9th layers) are still beneficial.    



\subsection{Discussion and Concluding Remarks}
While we show improvement in accuracy through feature manipulation, we highlight that we only explore straightforward boosting techniques. It is our hope that with further explorations, together with the other identified opportunities for confidence and localization adjustments, significant improvements in accuracy will be possible through the paradigm of collaborative machine intelligence.  

\section{Related Work}
Early efforts in describing the need for enabling collaborative intelligence among heterogeneous IoT devices, complementary to our vision, have been advocated recently~\cite{qiu2018kestrel, lee2018, hotmobile2019, eugene2019}. Qiu et al.\cite{qiu2018kestrel} describe a scenario where cameras of differing capabilities co-exist in the same network: fixed surveillance cameras and resource-constrained mobile devices with cameras. The authors demonstrate that moving vehicles can be tracked seamlessly across this heterogeneous camera network through selective actuation of devices without overly draining the mobile devices. In essence, the resource-intensive video analytics pipeline is performed on the cloud and the mobile cameras are consulted intermittently, only to resolve ambiguities. Further, Lee et al. \cite{lee2018} demonstrate significant savings in bandwidth needs (of \emph{dumb} cameras that offload raw footage to a central cloud) -- they show that by establishing space-time relationships, apriori, between co-existing cameras, that they can be selectively turned ON (and OFF) leading to as much as 238 times savings in bandwidth at a miss rate of only 15\% for a vehicle detection task. Similarly, Jain et al.~\cite{hotmobile2019} also show that significant correlations exist between co-located cameras, and discuss configurations of video analytics pipelines that can be triggered by peer cameras leading to both cost efficiency and superior inference accuracy. Unlike such past work, we focus explicitly on using collaboration to modify or abort the inferencing pipeline itself, instead of selectively activating nodes or performing fusion of the outputs from multiple nodes. Most recently, Yao et al.~\cite{eugene2019} describe the vision for providing machine intelligence as a service at the edge for resource-constrained devices. In addition to outlining core capabilities required for enabling such a service (e.g., scheduling, caching, resource profiling), they also describe opportunities for the convergence of the idea of collaboration between devices and deep intelligence as a service. Furthermore, Abdelzaher et al.~\cite{abdelzaher2020five} and Misra et al.~\cite{misra2019} outline the challenges and opportunities for machine intelligence at the edge. Most recently, in Hannaneh and Nadeem~\cite{hannaneh2019} put forth a similar vision for performing collaboration at the edge between multiple camera nodes, for gains in accuracy and  resource utilization, among other measures. Similar to our proposal, the authors propose that cameras can share feature summaries amongst them to improve the overall performance on vision-related tasks. While the authors share several novel ideas, in this paper, we identify multiple opportunities for collaborative gains, demonstrate an example of a collaborative operational mode that requires no re-training of the DNNs, and evaluate a working system of multiple peer cameras. 
\small
\bibliographystyle{abbrv}
\bibliography{main} 

\end{document}